%% file: iclr2027_conference_main.tex
\documentclass{article} 
\usepackage{iclr2027_conference,times}

\input{math_commands.tex}

\usepackage{hyperref}
\usepackage{url}

\usepackage{algorithm}
\usepackage{algorithmic}
\usepackage{graphicx}

\usepackage{booktabs}
\usepackage{multirow}
\PassOptionsToPackage{table}{xcolor}
\usepackage{amsfonts}
\usepackage{amsmath}

\usepackage[table,svgnames]{xcolor} 

\title{
GADAR: Generalizing Across Domains via Action Ranking for Classical Planning
}

\author{Rajesh Mangannavar, Pranay Dugar \& Prasad Tadepalli \\
Oregon State University\\
Corvallis, OR 97330, USA \\
\texttt{\{mangannr,dugarp,prasad.tadepalli\}@oregonstate.edu}
}

\newcommand{\gadar}{GADAR}                 
\newcommand{\gadarbind}{\gadar{}-BIND}     
\newcommand{\union}{UNION}                 
\newcommand{\todonum}[1]{#1}

\iclrfinalcopy 
\begin{document}

\maketitle

\begin{abstract}

Learned policies for classical planning act without search, but each is tied to the domain it was trained on: network modules per action schema, parameters per predicate, or vocabulary-sized encodings mean a trained policy cannot even be executed on a new domain's predicates and action schemas. We present \gadar{} (Generalizing Across Domains via Action Ranking), a policy that takes the domain description as input, $\pi(s, G, D)$, and constructs grounded actions in domains never seen in training, without search. Symbols are described by the roles they play in the domain rather than by their names, in three layers that each extend the previous one: (i) a \emph{lifted domain layer} embeds $D$ in every state graph through structural signatures; (ii) a \emph{binding layer} ties each ground action's applicability to the lifted preconditions it instantiates; and (iii) a \emph{chain layer} states which schema produces what another needs and how far each schema stands from the goal. The same network therefore runs on every domain, seen or unseen. We evaluate zero-shot generalization leave-one-domain-out over eight benchmark domains. A single \gadar{} policy trained on seven domains solves 58.3\% of their test instances, one point below the 59.3\% of dedicated single-domain \gadar{} models, so one policy serves many domains at no cost in coverage; a multi-task control that keeps symbol identity in its weights, while strong on some of the domains it trains on, transfers to none of the held-out ones. Zero-shot, \gadar{} executes on every held-out domain and exceeds random choice under the same executor on the large test instances of all eight: it solves 22.6\% of them on average, against 3.3\% for random choice and 0.6\% for the same representation without the chain layer, and reaches 68\% on held-out Visitall. Zero-shot coverage nonetheless remains well below in-domain coverage on most domains, and we analyze where it is weakest, including domains whose action schemas are structurally alike but play different roles.

\end{abstract}

\section{Introduction}

Classical planning finds action sequences achieving goals in deterministic
environments. Traditional planners solve small problems optimally by search
and heuristics but scale poorly \citep{complexity-planning, fast_downward},
which motivates learning-based planning: a model trained on solved problems
captures regularities that carry over to new ones. Two families of learned
guidance have developed largely in parallel.
Learned \emph{heuristics} estimate cost-to-go and guide a search algorithm
\citep{strips_hgn, karia2021learning, goose_v2}; learned \emph{policies}
select actions directly, executing plans greedily without search
\citep{toyer2020asnets, expressive_v1, gabar}.

Learned policies generalize well \emph{within} a domain. ASNets learn
per-schema network modules that transfer across problem sizes
\citep{toyer2020asnets}; relational GNN policies learn value functions whose
greedy execution solves instances far larger than those seen in training
\citep{expressive_v1, policy_beyond_c2}; GABAR encodes states as
action-centric graphs and constructs grounded actions by ranking them
\citep{gabar}. All share a design choice: the domain's schemas and
predicates are reflected in the network itself, as modules, parameters, or
feature dimensions. Shaping the network to its domain aids within-domain
generalization, but it is also what binds each policy to that domain, which
cannot even be executed where its schemas and predicates have no slots.

For learned heuristics, a different treatment of the domain has emerged.
STRIPS-HGN computes structural features that make no reference to symbol
identity \citep{strips_hgn}, and GOOSE's lifted learning graph encodes the
full domain description as part of the heuristic's input, training one
heuristic across many domains and evaluating on held-out ones
\citep{goose_v2}. The input side of learning for planning can therefore be
made domain-agnostic. Heuristics, however, delegate action construction to
search; the corresponding step for policies, which must construct grounded
actions themselves, remains open, and \citet{goose_v2} name it as future
work. It is not a trivial step: the same authors prove that message passing
over the lifted graph cannot compute even $h^{\mathrm{add}}$, and their
domain-independent setting is empirically their hardest.

In this work we take that step, and our key idea is to change what the
network is told about symbols: not which symbol something is, but what role
the symbol plays in the domain. A predicate is described by where it
appears (which schemas require it, which produce or destroy it, at which
argument positions), and under that description gripper's
$\textsf{pick}$-then-$\textsf{drop}$ and miconic's
$\textsf{board}$-then-$\textsf{depart}$ are one structure, which a policy
can learn in one domain and reuse in another. Two things are then needed to
make the description usable: every ground action's applicability must be
tied to that structure, so that ``why is this action available'' is said in
the domain's own vocabulary; and the relation planning turns on, that one
schema produces what another needs, must be stated explicitly rather than
left for message passing to recover. The result is a policy
$\pi(s, G, D)$ that is at once domain-general and search-free, where
classical planners buy generality with search and learned policies buy
reactivity with specialization. The question we study is what such a policy
can transfer, not whether it outraces search.

We build this capability as a ladder of three systems, each adding one
group of representational commitments to the previous one, so that every
adjacent comparison isolates that group. \union{} is the control: symbol-identity
features widened to the union of the training vocabularies,
which is multi-task learning with the domain in the weights. \gadarbind{}
replaces symbol identity with structural features and puts $D$ into every
state graph, as a lifted domain layer plus the \emph{binding layer}.
Structure alone is not yet transfer: ``schema $A$ produces what schema $B$
consumes'' is only a two-hop path through a shared predicate hub, and the
hub loses the pairing. Full \gadar{} therefore adds the \emph{chain
layer}, stating that relation directly and marking how far each schema
stands from the goal. All three share one encoder, one ranking decoder, and the
same search-free execution, so every comparison varies the representation
and not the architecture.

We evaluate zero-shot generalization leave-one-domain-out over all eight
benchmark domains. One \gadar{} policy serves every domain it is trained on:
trained on seven domains, it matches the in-domain coverage of dedicated
single-domain models (58.3\% against 59.3\%). Zero-shot, on the large test
instances of held-out domains, it exceeds random choice under the same
executor on all eight (22.6\% against 3.3\% on average), reaching 68\% on
Visitall at plan quality \todonum{0.97}, against 4\% without the chain layer
and 0\% for the union-vocabulary control. Elsewhere it solves 4--29\%, and
the per-domain results suggest where transfer breaks: when a held-out domain's schemas are structurally alike but
play different roles, as Logistics' truck and airplane actions do, features
that ignore symbol names may not tell them apart. \gadar{} is invariant by
construction to renaming of predicates, schemas, and objects, a robustness
current large reasoning models lack \citep{llm_baseline}; the same
invariance means that what distinguishes such schemas has to come from data
in the new domain.

\textbf{Contributions.}
\begin{itemize}
\item To our knowledge, the first domain-agnostic \emph{policy} for
  classical planning: $\pi(s, G, D)$ constructs grounded actions zero-shot
  in unseen domains, without search.
\item One policy for many domains: a single \gadar{} model trained on seven
  domains matches dedicated single-domain models in-domain on average, with
  positive transfer between some domains.
\item A graph representation with two novel relations, both computable in
  any domain: \emph{binding} edges, which link each applicable ground
  action's objects to the lifted precondition occurrences they satisfy and
  so express why the action is applicable in the vocabulary of $D$, and \emph{chain} edges,
  which express which schema produces or destroys what another needs and
  how far each schema is from the goal.
\end{itemize}

\section{Related Work}

\textbf{Learning heuristics across domains.} Domain-independent heuristics
have a long history; we focus on the neural line. STRIPS-HGN trains
hypergraph networks on structural features of the delete relaxation and
evaluates on unseen domains \citep{strips_hgn}. GOOSE introduces graph
representations of planning tasks for grounded and lifted inputs; its
lifted learning graph encodes the domain description itself (schemas,
preconditions, effects, argument bindings via index features), and its
domain-independent setting trains one heuristic across many domains and
evaluates on held-out ones \citep{goose_v1, goose_v2}. WL-GOOSE obtains
stronger per-domain heuristics from Weisfeiler--Lehman features, at the
cost of per-domain feature dictionaries \citep{wl_goose}.
\citet{karia2021learning} learn generalized heuristics over canonical
abstractions; LLM-evolved heuristic ensembles transfer to a held-out IPC
learning track \citep{llm_evolved_heuristics_2026}. These works establish
domain-agnostic inputs, and we adopt their lesson of structural features
and lifted domain graphs. They differ in output: a heuristic delegates
action construction to search, which stays in the execution loop.
\citet{goose_v2} further prove an expressiveness limitation of message
passing on their lifted graph and name lifted policy learning as open; our
binding layer adds exactly the information a policy needs and their graph
provably lacks (their Theorem 4.3; Appendix~\ref{app:llg}).

\textbf{Learned policies for planning.} Generalized policy learning aims at
policies that transfer across instances of a domain
\citep{approx_Policy_iter, relational_rl}. ASNets construct a network from
the domain's action schemas and ground atoms, with weight sharing across
instantiations of the same schema \citep{toyer2020asnets}. St{\aa}hlberg,
Bonet, and Geffner learn value functions with GNNs whose greedy policies
generalize across instance sizes, characterize their expressive power in
terms of two-variable logics, and extend the approach with policy gradients
and unsupervised objectives \citep{expressive_v1, unsupervised_gnn,
policy_gradients, policy_beyond_c2}. Relational deep RL approaches
similarly learn policies over relational state representations
\citep{relational_drl, sr_drl, relational_mdp_learning, rddl_gnn,
generalized_planning_deep_rl}. GABAR represents ground actions explicitly in the
state graph and constructs actions autoregressively by ranking
\citep{gabar}. The action-centric idea also extends to partial observability
\citep{gammazero, hoo_pomdp}, varying what the policy observes while
holding the domain fixed. The most recent work remains per-domain: self-improving
search-and-learn loops train a relational value heuristic per domain
\citep{learning_to_search_2026}, lookahead-based general policies are
learned over instances of a single domain
\citep{lookahead_policies_2026}, and Q-value policy learning is framed
per-domain explicitly \citep{mueller_qvalues}. In all of these the
vocabulary lives in the network as per-schema modules, per-predicate
parameters, or vocabulary-sized encodings, binding the policy to its
training domain. \gadar{} removes that binding by moving the domain
description from the network's parameters into its input.

\textbf{Ranking in planning.} Learning to rank has ordered states and
heuristic values for search \citep{ranking_old_2009, ranking_og,
chrestien_optimize, gbfs_v2}; GABAR instead ranks \emph{actions} locally
within a state \citep{gabar}, \citet{mangannavar2026scalable}. Cross-domain, this choice becomes
load-bearing: a shared value function presupposes a common scale of goal
distances that does not exist across domains, whereas local rankings are
scale-free. \citet{mueller_qvalues} report convergent evidence: Q-value
regression fails for lack of action discrimination and a ranking-style
margin repairs it.

\textbf{Large language models.} Large reasoning models degrade under symbol
renaming \citep{llm_baseline}, an invariance \gadar{} has by construction;
LLM pipelines that write or repair PDDL \citep{halo_orchestrator} leave
planning to search.

\section{Background and Problem Setup}

A classical planning problem $P = \langle D, I \rangle$ pairs a first-order
domain $D$ (predicate symbols $\mathcal{P}$ with arities, and action
schemas $\mathcal{A}$, each with parameters, preconditions $pre(a)$, and
add/delete effects $add(a)$, $del(a)$) with instance information
$I = \langle O, s_0, G \rangle$: objects, an initial state, and a goal
condition, the last two as sets of ground atoms. A ground action
substitutes objects for a schema's parameters, is applicable in $s$ if
$pre(a) \subseteq s$ under the substitution, and yields
$s' = (s \setminus del(a)) \cup add(a)$; a plan is an applicable sequence
from $s_0$ to a state satisfying $G$.

GABAR learns, per domain, a policy $\pi_D(s, G)$ that generalizes to larger
instances of the \emph{same} domain \citep{gabar}. We address the
cross-domain version: from solved small instances of training domains
$D_1, \dots, D_n$, learn a \emph{single} policy $\pi(s, G, D)$, with the
domain description itself an input, that solves instances of a test
domain $D_{\mathit{test}} \notin \{D_1, \dots, D_n\}$, zero-shot. Execution
is search-free: the highest-ranked applicable action is executed, a monitor optionally avoids revisiting states, and execution ends
at the goal, a dead end, or a length bound.

\begin{figure*}[t]
\centering
\includegraphics[width=0.98\textwidth]{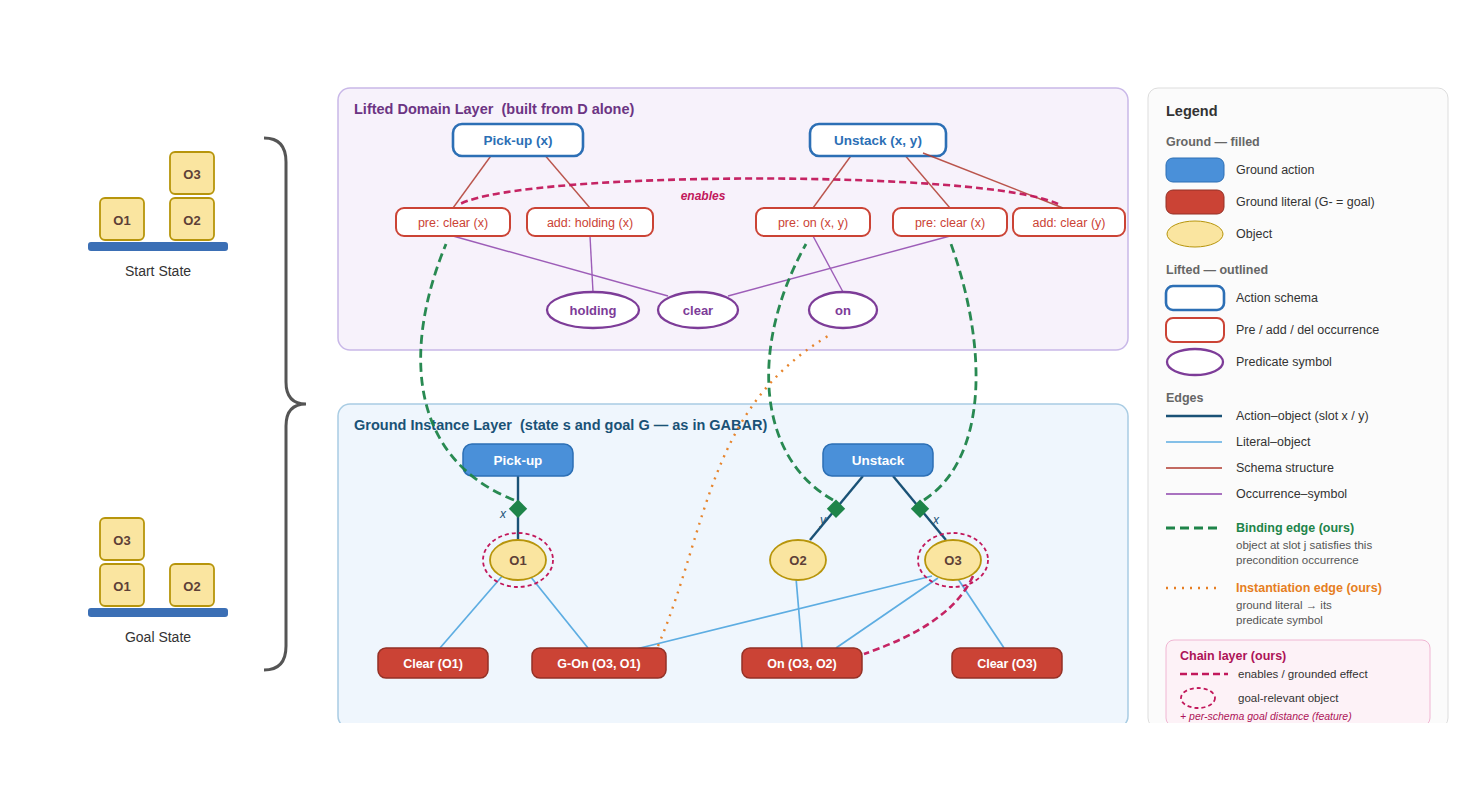}
\caption{The \gadar{} representation for a Blocksworld task. The
\emph{ground instance layer} (bottom) is the action-centric graph of the
grounded policy we build on \citep{gabar}: objects, ground literals of the state and goal (G- prefix), and ground
action nodes with action--object edges carrying parameter slots. The
\emph{lifted domain layer} (top) is built from $D$ alone: schemas, their
precondition and effect occurrences, and predicate symbols. \emph{Binding
edges} (green, dashed) link an object's slot on an applicable ground action
to the precondition occurrences it satisfies; \emph{instantiation edges}
(orange, dotted) link each ground literal to its predicate symbol. Filled
nodes are ground, outlined nodes lifted. The chain layer (magenta) is
shown by one \emph{enables} edge, one grounded deletes-true edge, and
goal-relevance rings.}
\label{fig:representation}
\end{figure*}

\begin{figure*}[t]
\centering
\includegraphics[width=1.0\textwidth]{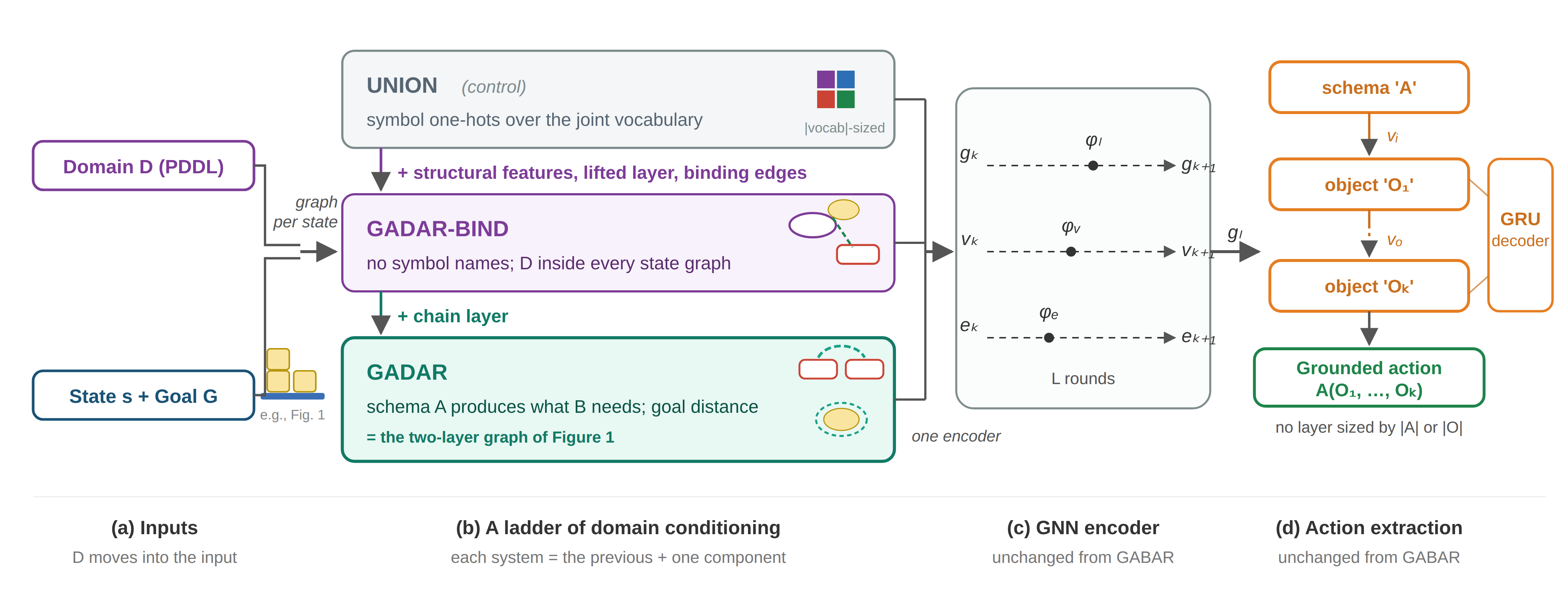}
\caption{\gadar{} overview. (a) $D$, state, and goal become one graph.
(b) The ladder: \union{} (symbol one-hots), \gadarbind{} (structural
features, lifted and binding layers), \gadar{} (chain layer). (c) GNN
encoder. (d) GRU decoder: schema, then arguments, scored against node
embeddings; nothing is sized by the number of schemas or objects.}
\label{fig:system}
\end{figure*}

\section{GADAR}

Given a planning instance in PDDL format together with its domain
description $D$, \gadar{} converts the pair into a graph, processes the
graph with a GNN encoder, and constructs a grounded action with a GRU-based
decoder that first selects an action schema and then each of its arguments
(Figure~\ref{fig:system}). Encoder and decoder are those of the per-domain
policy of \citet{gabar}; what changes is
the graph they operate on. We present the representation as it is
evaluated: a ladder of systems in which each addition is measured against
its predecessor.

\textbf{Encoder.} Features are embedded into $d{=}64$ dimensions, then
$L{=}9$ rounds update edges, nodes (attention over incoming edges), and a
global node, with skip connections to the input encodings
(Appendix~\ref{app:encoder}). The output is one embedding per node, which
the decoder ranks.

\subsection{A Domain-Agnostic Graph Representation}

\textbf{Ground instance layer.} The base of the representation is that
action-centric graph: nodes for objects, for ground literals
of the current state and the goal (goal literals marked by a flag), and for
action schemas, together with a global node; edges connect literals to their
argument objects and schemas to the objects of their applicable groundings,
with parameter positions on the edges (Figure~\ref{fig:representation},
bottom). This graph is what makes search-free action construction possible:
the decoder selects among schema nodes and object nodes that are present in
the input.

\textbf{The graph.} A task $(s, G, D)$ is encoded as
$\mathcal{G} = (V, E, X, R)$. The node set is
\begin{equation*}
V \;=\; O \,\cup\, L \,\cup\, A \,\cup\, \Sigma \,\cup\, Q \,\cup\, \{g\},
\end{equation*}
where $O$ are the objects, $L$ the ground literals of the state and goal,
$A$ the action schemas (one node per schema), $\Sigma$ the
predicate symbols of $D$, including one synthetic unary static predicate
per declared type so that typed and predicate-typed domains present the
same structure, $Q$ the \emph{occurrence nodes} $q_{a,k}$, one per
literal occurrence $k$ in the preconditions and effects of schema $a$, and
$g$ a global node. The edge set is a union of typed relations,
\begin{equation*}
E \;=\; E_{\mathrm{lit}} \cup E_{\mathrm{act}} \cup E_{\mathrm{inst}} \cup
E_{\mathrm{occ}} \cup E_{\mathrm{bind}} \cup E_{\mathrm{chain}} \cup
E_{\mathrm{eff}},
\end{equation*}
\begin{itemize}
\item $E_{\mathrm{lit}} \subseteq L \times O$, $E_{\mathrm{act}} \subseteq
  A \times O$: literals to their arguments and schemas to the objects of
  their \emph{applicable} groundings (the ground layer), with argument
  and slot positions as edge features;
\item $E_{\mathrm{inst}} \subseteq (L \cup O) \times \Sigma$:
  \emph{instantiation}, each literal to its predicate's symbol and each
  object to its declared type's symbol;
\item $E_{\mathrm{occ}} \subseteq Q \times (\Sigma \cup A)$: each occurrence
  to its predicate symbol and to its schema, carrying the role
  $r(q) \in \{\mathrm{pre}, \mathrm{add}, \mathrm{del}\}$ and the
  argument-to-slot pairs $(i, j)$ of that occurrence;
\item $E_{\mathrm{bind}} \subseteq O \times Q$, the \emph{binding layer}:
  $(o, q_{a,k})$ whenever some applicable grounding of $a$ places $o$ at a
  slot that occurrence $k$ constrains;
\item $E_{\mathrm{chain}} \subseteq Q \times Q$, the \emph{chain layer}:
  $\textit{enables} = \{(q, q') : r(q) = \mathrm{add},\, r(q') =
  \mathrm{pre},\, \mathrm{pred}(q) = \mathrm{pred}(q')\}$ and
  \textit{threatens} likewise with $r(q) = \mathrm{del}$;
\item $E_{\mathrm{eff}} \subseteq O \times L$, the grounded effect edges: for
  each applicable grounding, an instantiated add effect equal to an
  \emph{unsatisfied} goal atom $\ell$ (\textit{achieves-goal}), or a delete
  effect equal to a \emph{true} atom $\ell$ (\textit{deletes-true}), links
  the grounding's objects to $\ell$.
\end{itemize}
Node features $X : V \to \{0,1\}^{d}$ concatenate indicator segments: the
node's kind; the arity (clamped at global caps $k_p$, $k_a$) and a
static-predicate flag; a goal flag on literals; on schema nodes the
goal-distance one-hot $\mathrm{gdist}(a) \in \{0, 1, 2, 3{+}\}$, the BFS
depth over \textit{enables} from the schemas that add a goal predicate; and
on objects a goal-relevance bit (the object occurs in an open goal atom).
Edge features $R : E \to \{0,1\}^{k}$ concatenate the relation type with
position, slot, and role indicators up to the caps. The widths $d$ and $k$
are functions of $(k_p, k_a)$ alone; no dimension is $|A|$, $|P|$,
$|T|$, or $|O|$. A single network therefore serves every domain, instances from
different domains batch together, and permuting symbol names leaves
$(X, R)$ unchanged: renaming invariance holds by construction. Every
feature is a flag or a one-hot; nothing is a magnitude. The goal distance
in particular is bucketed rather than numeric, deliberately: a number must
be read on a scale, and no scale is shared across domains. The same
argument makes local action ranking, rather than value regression, the
right objective in this setting.

\textbf{Cost.} $\Sigma$, $Q$, $E_{\mathrm{occ}}$, and $E_{\mathrm{chain}}$
are domain-sized constants computed once per domain; $\mathrm{gdist}$
depends only on which predicates occur in the goal and is computed once per
(domain, goal signature). Only $E_{\mathrm{bind}}$, $E_{\mathrm{eff}}$, and
the goal marks are per-state, each linear in the applicable groundings and
open goal atoms.

\textbf{Why these relations.} $E_{\mathrm{bind}}$ connects what the two
prior lines capture only separately: LLG has lifted preconditions but no
ground actions, and GABAR has ground applicability but indexed by
per-domain strings that do not transfer. Binding expresses why each action is applicable in $D$'s own
vocabulary.\footnote{Binding edges are \emph{cross-layer}: each joins a
node of the ground instance layer (an object of the current problem) to one
of the lifted domain layer (a precondition occurrence of a schema in $D$).
The graph contains both layers; the binding edges connect them.} In
Figure~\ref{fig:representation}, the edge placing O3 in Unstack's first
slot binds to the occurrences \emph{on(x,y)} and \emph{clear(x)}, and this
information survives unchanged if every symbol is renamed.
$E_{\mathrm{chain}}$ states directly that schema $A$ produces what schema
$B$ needs. Without it, this relation exists only as a two-hop path through
a shared predicate hub, and the hub's aggregation destroys the
producer--consumer pairing. Because the chain edges connect occurrences,
the pairing survives message passing, delete effects acquire a consequence,
and the relation transfers across domains: gripper's
$\textsf{pick} \rightarrow \textsf{drop}$ and miconic's
$\textsf{board} \rightarrow \textsf{depart}$ are the same structure, an
\textit{enables} edge between an add- and a pre-occurrence of a shared
predicate. $E_{\mathrm{eff}}$ grounds the layer: the decoder scores object
nodes, and these edges place ``this object closes an open goal'' and ``this
object undoes something true'' one hop from the quantity being ranked.

\subsection{A Ladder of Domain Conditioning}

The three systems stand in a strict subset relation. \union{} (control)
keeps the symbol-identity one-hots, widened to the union of the training
vocabularies: the domain stays in the weights, which is multi-task
learning, not domain conditioning. \gadarbind{} replaces those one-hots
with the structural features $X$, making the policy executable in any
domain, and adds $\Sigma$, $Q$, $E_{\mathrm{inst}}$, $E_{\mathrm{occ}}$,
and $E_{\mathrm{bind}}$. \gadar{} adds $E_{\mathrm{chain}}$,
$E_{\mathrm{eff}}$, and the goal-distance and goal-relevance features.
Because adjacent systems differ in one group of components, the experimental
comparison and the ablation study are the same object: the
\union{}-to-\gadarbind{} step measures what removing symbol identity costs
and what the domain description buys, and the \gadarbind{}-to-\gadar{}
step measures the chain layer together with the grounded effect edges and
goal features it adds. We do not separate those three here, and attribute
the second step to them jointly.

\subsection{Action Construction and Training}

The decoder changes only where per-domain constants would otherwise appear
(Figure~\ref{fig:system}d). A GRU selects a schema by
scoring schema-node embeddings against its hidden state, then each argument
in turn by scoring object-node embeddings, feeding every selection back as
the next input. The argument loop runs to the selected schema's arity, read
from $D$ under a global arity cap rather than a per-domain constant. No
component is sized by the number of schemas, objects, or predicates;
selection is always a dot product between the decoder state and the
embedding of a node present in the graph, which is what lets one network
construct actions in a domain whose schemas it has never seen. Training is
supervised: cross-entropy against the expert's schema and argument
choices at every state along planner-generated plans. Because
feature widths are domain-independent, domains mix freely within a
batch.

\section{Experimental Setup}

\textbf{Domains.} We evaluate on the eight benchmark domains used by GABAR:
Blocks, Miconic, Spanner, Gripper, Visitall, Grid, Logistics, and
Rovers \citep{gabar, pddl-generators}. Training data consists of plans
computed by Fast Downward with the LAMA-first satisficing configuration
\citep{fast_downward} on the small instances of each training domain
(between 125 and 312 instances per domain); states along each plan are
converted to graphs and the policy is trained to reproduce the expert
action as described above. Each domain in turn is held out and evaluated
on two splits.

\textbf{Protocol.} Two experiment families, run for every rung of the
ladder. (i)~\emph{Zero-shot generalization} (leave-one-domain-out, the
main experiment): for each of the eight domains, a policy is trained on the remaining seven and evaluated
on the held-out domain zero-shot, having never seen its predicates,
schemas, or objects. (ii)~\emph{In-domain, multi-domain}: the same
seven-domain policies
are evaluated on the domains they trained on, and for each domain we report
the mean over the seven policies whose training set includes it; a
single-domain variant, \gadar{} (single-domain), trains the same representation on that
domain alone. The \emph{test} split is the domain's test
set, with instances up to an order of magnitude larger than training-scale
problems; the \emph{training-scale} split is the held-out domain's own
training-scale instances, unseen because the domain was held out, which
separates failing on the unseen domain from failing to scale within it.
Training domains are evaluated on the test split only.

\textbf{Baselines and references.} Our primary comparison is the \union{}
control described above: identical data, architecture, and training
procedure, with only the representation different. Any zero-shot
capability \gadar{} has beyond this control comes from moving the domain
into the input. Two floors bound the zero-shot numbers from below. The
\emph{random floor} is the same executor and monitor choosing uniformly
among applicable ground actions, averaged over three rollouts per problem;
it measures how far legal moves alone get on each split. The
\emph{untrained control} is the identical system, with the same graph,
decoder, executor, and step bound, but a network that has seen a single
epoch of training. As a reference for dedicated training we report
published per-domain GABAR \citep{gabar}, trained directly on each
held-out domain and evaluated on its own test instances, averaged over its
three published size bands. Those instances are not identical to ours, so
GABAR (published) indicates what per-domain training achieves at this scale
rather than serving as a matched comparison.

\textbf{Ablations and oracle.} The ladder is the ablation study
(Figure~\ref{fig:system}b), and each group of components is measured by
adjacent columns of Table~\ref{table:main_table}. We additionally report
one diagnostic that is not a rung of the ladder: an \emph{achievement
oracle} that, at every state, prefers any applicable ground action whose
add effects directly satisfy an unsatisfied goal atom, and otherwise defers
to the policy. It bounds what a hand-coded terminal trigger can achieve on
each domain.

\textbf{Metrics and implementation.} We report coverage (C), the
percentage of test problems solved within a 500-step bound, and
plan quality (P), the ratio of the reference planner's plan length to the
policy's on jointly solved problems, so values above 1 mean plans shorter
than the reference; coverage is the primary metric
(Appendix~\ref{app:metrics}). Hyperparameters are carried over from
\citet{gabar} unchanged (Appendix~\ref{app:hyper}), and renaming invariance
is verified by a unit test (Appendix~\ref{app:invariance}). In-domain
results use the checkpoint with the lowest combined training and validation
loss; zero-shot results use the training epoch that maximizes mean
zero-shot coverage on the other held-out domains.

\section{Results and Discussion}

Table~\ref{table:main_table} presents the zero-shot results for all eight
held-out domains and Table~\ref{table:indomain} the in-domain results. We
organize the discussion so that each question is answered by one
comparison: a rung of the ladder, a control, or a diagnostic.

\begin{table}[t]
\centering
\small
\setlength{\tabcolsep}{3pt}
\caption{Zero-shot generalization, leave-one-domain-out: coverage (\%) on
each held-out domain, which is never seen in training. Training-scale: its
small instances; Test: its test instances. Rand: uniform choice among
applicable actions under the same executor. Mean: over the eight held-out
domains.}
\label{table:main_table}
\begin{tabular*}{\linewidth}{@{\extracolsep{\fill}}ll|cccccccc|c@{}}
\toprule
& Held out & Blocks & Gripper & Miconic & Visitall & Grid & Logistics & Spanner & Rovers & Mean \\
\midrule
\multirow{4}{*}{\shortstack[l]{Training-\\scale}} & Rand & 1.7 & 0.0 & 88.3 & 99.7 & 3.0 & 0.0 & 23.4 & 15.2 & 28.9 \\
& \union{} & 0.0 & 0.0 & 0.0 & 0.8 & 1.0 & 0.0 & 0.0 & 0.0 & 0.2 \\
& \gadarbind{} & 2.5 & 0.0 & 0.0 & 16.0 & 3.1 & 0.0 & 0.0 & 0.0 & 2.7 \\
& \textbf{\gadar{}} & 37.5 & 13.3 & 36.3 & 100.0 & 33.4 & 9.9 & 43.4 & 16.1 & 36.2 \\
\midrule
\multirow{4}{*}{Test} & Rand & 0.2 & 0.0 & 1.1 & 6.0 & 0.0 & 0.0 & 13.8 & 5.2 & 3.3 \\
& \union{} & 0.0 & 0.0 & 0.0 & 0.0 & 0.0 & 0.0 & 0.0 & 0.0 & 0.0 \\
& \gadarbind{} & 1.0 & 0.0 & 0.0 & 4.0 & 0.0 & 0.0 & 0.0 & 0.0 & 0.6 \\
& \textbf{\gadar{}} & 23.0 & 6.0 & 19.0 & 68.0 & 12.1 & 4.0 & 29.2 & 19.6 & 22.6 \\
\bottomrule
\end{tabular*}
\end{table}

\begin{table}[t]
\centering
\small
\setlength{\tabcolsep}{3pt}
\caption{In-domain coverage (\%) on the test instances. Ladder rows average
the seven leave-one-out policies whose training set includes the domain;
\gadar{} (single-domain) is trained on that domain alone. GABAR (published):
per-domain GABAR as published, on its own test instances.}
\label{table:indomain}
\begin{tabular*}{\linewidth}{@{\extracolsep{\fill}}l|cccccccc|c@{}}
\toprule
& Blocks & Gripper & Miconic & Visitall & Grid & Logistics & Spanner & Rovers & Mean \\
\midrule
GABAR (published) & 100.0 & 100.0 & 100.0 & 90.7 & 96.3 & 79.0 & 92.0 & 82.0 & 92.5 \\
\midrule
\union{} & 57.2 & 84.1 & 86.1 & 77.3 & 89.6 & 18.6 & 48.6 & 9.6 & 58.9 \\
\gadarbind{} & 43.4 & 95.4 & 67.0 & 78.6 & 64.9 & 0.3 & 40.0 & 8.2 & 49.7 \\
\gadar{} & 74.6 & 94.5 & 61.9 & 81.1 & 94.6 & 2.7 & 47.9 & 9.0 & 58.3 \\
\midrule
\gadar{} (single-domain) & 95.5 & 100.0 & 41.2 & 54.0 & 87.5 & 3.1 & 85.4 & 7.4 & 59.3 \\
\bottomrule
\end{tabular*}
\end{table}

\textbf{Is the representation costly on its own? (\gadar{} (single-domain) vs.\ GABAR (published).)}
Trained on a single domain, \gadar{} (single-domain) is within five
points on Blocks and Gripper and within ten on Spanner and Grid, but far
behind on Miconic and Visitall, and it reaches only 3.1\% and 7.4\% on
Logistics and Rovers. Describing symbols by role rather than name thus
costs little in four domains and much in the others; on Logistics,
\union{}, which keeps the names, reaches 18.6\% (see the aliasing
discussed below).

\textbf{Does one model serve several domains? (\gadar{} vs.\ \gadar{} (single-domain).)}
Every ladder column is a policy that serves seven domains at once. Averaged
over the seven such policies that train on each domain, \gadar{} solves
58.3\% of the test instances, against 59.3\% for the eight
\gadar{} (single-domain) models: sharing one model across seven domains costs one point
on average. The average hides transfer in both directions. Training
alongside other domains helps Visitall (81.1\% against 54.0\%), Miconic
(61.9\% against 41.2\%), and Grid (94.6\% against 87.5\%), and hurts Spanner
(47.9\% against 85.4\%) and Blocks (74.6\% against 95.5\%). Logistics and
Rovers are equally low either way, so their gap to GABAR (published) is a
cost of the representation, not of sharing.

\textbf{Does the domain need to be in the input? (\union{} vs.\ \gadar{}.)}
\union{}, trained on the identical data with the identical architecture,
is the stronger model on some training domains (86.1\% on in-domain Miconic
and 18.6\% on Logistics, against \gadar{}'s 61.9\% and 2.7\%) and fails
completely off them: it solves 0\% of every held-out test split, and its
top-ranked proposal is applicable in fewer than \todonum{1}\% of states
there. On the same domains \gadar{} solves 4.0--68.0\%. A
symbol-identity representation cannot act on a vocabulary it has not seen;
moving the domain description into the input is what makes zero-shot
execution possible at all. Averaged in-domain, \union{} reaches 58.9\%
against \gadar{}'s 58.3\%: giving up symbol names costs under a point on
average, but it costs Miconic and Logistics.

\textbf{Is the description enough? (\gadarbind{} vs.\ \gadar{}.)}
\gadarbind{} already contains the full domain description and the binding
edges, yet it solves 0.6\% of held-out test instances on average and none on six
of the eight domains. The chain layer, together with the grounded effect edges and goal features
added in the same step, raises test coverage on every held-out domain: to
22.6\% on average, and from 4.0\% to 68.0\% on Visitall. That step
introduces no new supervision and no domain-sized parameters: its relations
are restated from $D$, and it widens only the domain-independent
edge-feature input. The description alone does not transfer; it must be
restated in a form message passing can use.

\textbf{Is it the policy or the executor? (Floors vs.\ \gadar{}.)} The
executor resamples an applicable action when the rollout re-enters a
visited state, so part of any coverage may come from the executor rather
than the policy. Two floors separate them. On small instances the executor
alone often suffices: random choice among applicable actions under the same
executor solves 99.7\% of training-scale Visitall and 88.3\% of
training-scale Miconic, so the training-scale split can credit the policy
with little there. The test split can: random choice solves 3.3\% of it on
average and at most 13.8\% (Spanner), and \gadar{} exceeds it on every
held-out domain, by 4.0 (Logistics) to 62.0 points (Visitall).
The two floors also differ from each other, and the difference matters: an
untrained network is not a neutral policy but a consistently biased one,
and it falls below random wherever random exploration is enough
(\todonum{2.6}\% against 88.3\% on training-scale Miconic,
\todonum{0.0}\% against 6.0\% on test Visitall). A policy should therefore
be credited against the higher of the two floors, which wherever we
measured both is random choice. The same effect explains the one split where
\gadar{} falls below random choice, training-scale Miconic (36.3\% against
88.3\%); on test Miconic, where exploration no longer suffices (1.1\%), it
solves 19.0\%.

\textbf{Are the plans goal-directed? (Plan quality.)} Where random choice
also succeeds, plan length separates a goal-directed policy from legal
exploration. On training-scale Visitall, \gadar{}'s plans average
\todonum{36.3} steps against 29.2 for the planner, a plan quality of
\todonum{0.81}; random choice solves nearly the same instances with plans
averaging 107.1 steps, a plan quality of 0.27. On the test split, where
random choice rarely succeeds, \gadar{}'s plan quality is \todonum{0.97}:
its plans are within \todonum{3}\% of the planner's, which searches while
\gadar{} does not.

\textbf{Is it a terminal trigger? (Achievement oracle vs.\ \gadar{}.)} A
policy might only learn to fire when a goal atom is one action away. The
achievement oracle hard-codes exactly that behavior and defers to the
policy otherwise. On test Visitall it solves \todonum{--}\% against
\gadar{}'s 68.0\%, so the policy is also choosing actions several
steps from achieving anything, which is what the bucketed goal distance
over the enables relation was designed to support.

\textbf{How much of in-domain coverage does zero-shot recover? (Table~\ref{table:main_table} vs.\ Table~\ref{table:indomain}.)}
Averaged over the eight domains, zero-shot \gadar{} solves 22.6\% of the
test instances against 58.3\% for \gadar{} policies that train on the
domain, about two fifths. The fraction ranges from 84\% on Visitall to
under a seventh on Grid and Gripper. On Logistics and Rovers zero-shot
matches or exceeds in-domain coverage: training on these domains does not
help either, which points to a representational limit rather than a
failure of transfer.

\textbf{Where is zero-shot generalization weakest? (Logistics and Gripper.)}
Logistics and Gripper are the weakest held-out domains, at 4.0\% and 6.0\%
of their test instances. Logistics illustrates one cause. Its two transport
modes use schemas, \textsf{load-truck} and \textsf{load-airplane} and their
unload counterparts, whose lifted structure is identical up to the vehicle
type they bind, so a representation that ignores symbol names gives them
the same features even though only one advances the goal in a given state.
Renaming invariance equates whole domains with identical structure; schemas
that are structurally alike \emph{within} a domain can still play different
roles in it. The in-domain results point the same way: with Logistics among
its training domains, \gadar{} still reaches only 2.7\% on it, and 3.1\%
when trained on Logistics alone, against 94.6\% on Grid. Gripper shows that
shared structure is not sufficient either: it shares Miconic's
producer--consumer pattern yet transfers less (6.0\% against 19.0\%). Whether structural similarity, between schemas within a
domain or between domains, predicts zero-shot coverage across all eight
domains we leave to a systematic analysis.

\section{Conclusion}

We introduced \gadar{}, a policy for classical planning that takes the
domain description as input and constructs grounded actions in domains
never seen in training, without search. The capability rests on the
binding layer, which connects ground applicability to lifted structure,
and the chain layer, which makes the domain's producer--consumer structure
and its distance to the goal explicit; neither adds supervision or
domain-sized parameters. One \gadar{} policy serves seven domains at once at
the in-domain coverage of dedicated single-domain models (58.3\% against
59.3\%). Zero-shot, the chain layer raises held-out test coverage from 0.6\% to
22.6\% on average, above random choice on all eight domains and to 68\% on
Visitall, though it remains well below in-domain coverage on most. Schemas that are structurally
alike but play different roles, as in Logistics, may mark a limit that
renaming-invariant features cannot cross without data from the new domain.
Few-shot adaptation, lifted successor generation, richer PDDL fragments,
and probabilistic settings remain future work.

\bibliography{iclr2027_conference}
\bibliographystyle{iclr2027_conference}

\appendix
\section{Appendix}

This appendix gives the detail a reader needs to implement the
representation, reproduce the experiments, or reuse either in another
setting. It adds no results: everything reported in the paper is in the
paper.

\subsection{Building the Graph}
\label{app:pseudocode}

Algorithm~\ref{alg:graph} constructs $\mathcal{G} = (V, E, X, R)$ for one
task, in the notation of the representation section. Lines marked $\dagger$
depend on the domain description alone, so they are computed once per domain
and reused for every state of it; the rest are per state and are linear in
the applicable groundings and the open goal atoms. The three systems are
prefixes of the same procedure: \union{} runs lines 1 to 3 with
symbol-identity features in place of $X$, \gadarbind{} runs lines 1 to 7, and
\gadar{} runs all of it.

\begin{algorithm}[t]
\caption{Construction of the \gadar{} graph for $(s, G, D)$}
\label{alg:graph}
\begin{algorithmic}[1]
\STATE $V \leftarrow O \cup L \cup A \cup \{g\}$ where $L$ holds the ground
  literals of $s$ and of $G$, the latter flagged as goal atoms
\STATE add $E_{\mathrm{lit}}$: each literal to its argument objects, argument
  position on the edge
\STATE add $E_{\mathrm{act}}$: each schema to the objects of its
  \emph{applicable} groundings, parameter slot on the edge
\STATE $\dagger$ compile each declared PDDL type of $D$ into a unary static
  predicate, so typed and predicate-typed domains present one structure
\STATE $\dagger$ $V \leftarrow V \cup \Sigma \cup Q$: a node per predicate
  symbol of $D$, and an occurrence node $q_{a,k}$ per literal occurrence in
  the preconditions and effects of each schema $a$
\STATE $\dagger$ add $E_{\mathrm{occ}}$: each occurrence to its predicate
  symbol and its schema, carrying role $r(q)$ and argument-to-slot pairs
\STATE add $E_{\mathrm{inst}}$ and $E_{\mathrm{bind}}$: each literal to its
  predicate symbol, each object to its type symbol, and $(o, q_{a,k})$
  whenever an applicable grounding of $a$ puts $o$ at a slot that occurrence
  $k$ constrains
\STATE $\dagger$ add $E_{\mathrm{chain}}$: \textit{enables} between an add and
  a pre occurrence of a shared predicate, \textit{threatens} likewise for del
\STATE add $E_{\mathrm{eff}}$: from each applicable grounding, link its
  objects to any unsatisfied goal atom it adds and any true atom it deletes
\STATE compute $\mathrm{gdist}(a)$ by breadth-first search over
  \textit{enables} from the schemas adding a goal predicate; mark objects
  occurring in an open goal atom
\STATE emit $X$ and $R$ as indicator vectors; widths depend only on the global
  arity caps $(k_p, k_a)$, never on $|A|$, $|P|$, $|T|$ or $|O|$
\end{algorithmic}
\end{algorithm}

Two implementation points matter for anyone rebuilding this. First, the type
compilation on line 4 is what lets one model span domains that declare PDDL
types and domains that spell types as ordinary unary predicates; without it
the two families present different structure for the same constraint, and a
policy trained on one produces type-invalid groundings on the other. Second,
$E_{\mathrm{bind}}$ on line 7 is the only relation that crosses between the
two layers, joining an object of the current problem to a precondition
occurrence of a schema in $D$. Everything else lives wholly inside one layer.

The reference implementation accumulates edge features sparsely, as a map
from node pair to the set of features that hold. A dense
$(k_a, |V|, |V|, \dim R)$ array is the obvious alternative and is what we
wrote first, but it costs $O(k_a |V|^2 \dim R)$ to allocate and scan for
every state, which dominates the cost of re-featurizing states during
evaluation on large instances.

\subsection{Encoder}
\label{app:encoder}

Node, edge, and global features are first embedded by
separate MLPs into $d{=}64$ dimensions. Each of $L{=}9$ rounds then (1)
updates every edge from its two endpoints, its own embedding, and the global
node, (2) updates every node from its incoming edges, aggregated by
attention (a softmax over the node's incoming edges), and (3) updates the
global node from all nodes and edges. At every round the current embeddings
are concatenated with their round-0 encodings before the update, a skip
connection that keeps the input features visible at depth. The output is
one embedding per node, which the decoder ranks.

\subsection{Relation to GOOSE's Lifted Learning Graph}
\label{app:llg}

LLG and our lifted layer share the essential idea of the domain
description as input, and differ in encoding: argument positions as edge
features rather than chains of index nodes, and occurrence nodes carrying
exact argument-to-slot pairs. The fundamental difference is purpose. A
heuristic graph must support one number read from a global embedding, so
LLG contains no ground actions and no applicability information, and
\citet{goose_v2} prove message passing on it cannot compute
$h^{\mathrm{add}}$. A policy graph must contain the decoder's targets, the
schema and object nodes to score, and the evidence for scoring them: which
groundings are applicable and why. The binding layer supplies that evidence
in $D$'s own vocabulary.

\subsection{Decoding and Training}
\label{app:decode}

A GRU scores the schema nodes against its hidden state and takes the best,
then scores the object nodes once per argument of the selected schema,
feeding each choice back as the next input. The number of arguments is read
from $D$ for the selected schema, under a global arity cap rather than a
per-domain constant, which is what allows the loop to run in a domain whose
schemas were never seen. Selection is a dot product between the decoder state
and the embedding of a node present in the graph, so no layer is sized by the
number of schemas, objects, or predicates. Search over these choices is a
beam of width 2 over schemas and 3 over objects per slot.

Training is supervised on plans from a satisficing planner: cross-entropy
against the planner's schema choice and each of its argument choices, at
every state along every plan. Because feature widths do not depend on the
domain, states from different domains mix freely within a batch.

\subsection{Domains and Data}
\label{app:data}

We use the eight benchmark domains of \citet{gabar}, generated with
\citet{pddl-generators}. Table~\ref{tab:data} gives the sizes. Instances the
planner fails to solve within its limit are skipped, so the collected count
can fall below the number of instances the generator produced; the counts in
the table are what was actually collected.

\begin{table}[t]
\small
\centering
\caption{Training data per domain: the small instances collected, and the
labelled states taken from their plans.}
\label{tab:data}
\begin{tabular}{@{}lrr@{}}
\toprule
Domain & Problems & States \\
\midrule
Blocks     & 200 & 4{,}216 \\
Gripper    & 147 & 7{,}287 \\
Miconic    & 228 & 5{,}453 \\
Visitall   & 125 & 3{,}654 \\
Grid       & 192 & 6{,}880 \\
Logistics  & 156 & 6{,}732 \\
Spanner    & 234 & 3{,}801 \\
Rovers     & 312 & 3{,}001 \\
\midrule
Total      & 1{,}594 & 41{,}024 \\
\bottomrule
\end{tabular}
\end{table}

Ten percent of each domain's states are held out for validation, stratified
by domain so every training domain is represented, and the split is
deterministic given the seed. In the leave-one-domain-out experiments the
held-out domain contributes nothing to either split: its plans are never
collected for that run, so no state, predicate, or schema of it is seen.
Testing is done on 300 problems each for all domains (combined from easy, medium and hard problems from the GABAR experiments). 

\subsection{Hyperparameters}
\label{app:hyper}

Table~\ref{tab:hyper} lists every setting. Encoder and decoder
hyperparameters are inherited from \citet{gabar} without modification. This
is deliberate rather than incidental: the comparisons in the paper are
between representations, and re-tuning the network for each one would
confound them. We therefore ran no architecture search, and a practitioner
adapting this to a new benchmark should expect to gain from tuning we did not
do. The two settings we did choose are the global arity caps $(k_p, k_a)$,
set to the maxima over the training domains so a held-out domain with wider
symbols still featurizes at full width, and the number of training epochs.

\begin{table}[t]
\small
\centering
\caption{Final hyperparameters, identical across all three systems except
where noted.}
\label{tab:hyper}
\begin{tabular}{@{}lr@{}}
\toprule
Setting & Value \\
\midrule
\multicolumn{2}{@{}l}{\emph{Encoder}} \\
Message-passing rounds $L$        & 9 \\
Representation size               & 64 \\
Attention heads                   & 1 \\
MLP layers per update block       & 2 \\
Global node                       & enabled \\
\midrule
\multicolumn{2}{@{}l}{\emph{Decoder}} \\
GRU layers                        & 3 \\
Beam width, schemas               & 2 \\
Beam width, objects per slot      & 3 \\
\midrule
\multicolumn{2}{@{}l}{\emph{Optimization}} \\
Optimizer                         & Adam \\
Learning rate                     & $5 \times 10^{-4}$ \\
Batch size                        & 64 \\
Dropout                           & 0.0 \\
Attention dropout                 & 0.0 \\
Weight decay                      & 0.0 \\
Epochs, structural systems        & 500 \\
Epochs, \union{}                  & 100 \\
\midrule
\multicolumn{2}{@{}l}{\emph{Evaluation}} \\
Execution bound (steps)           & 500 \\
Search at test time               & none (greedy) \\
Revisit monitor                   & enabled \\
Training plan generator           & \texttt{fd-lama-first} \\
Plan-quality reference            & \texttt{fd-lama-first} \\
\bottomrule
\end{tabular}
\end{table}

A run is seeded once, and that seed governs the network initialization, the
training shuffle, and the train and validation split; the revisit monitor's
fallback sampling is seeded per problem, so that it is identical across
execution harnesses. Each configuration is a single run.

\subsection{Evaluation Protocol}
\label{app:metrics}

Let $\mathcal{T}$ be a test set and $\mathrm{solved}(\pi) \subseteq
\mathcal{T}$ the instances on which policy $\pi$ reaches a goal state within
the execution bound. \emph{Coverage} is
\begin{equation*}
C(\pi) \;=\; \frac{|\mathrm{solved}(\pi)|}{|\mathcal{T}|} \times 100 .
\end{equation*}
Execution is greedy and search-free. At each state the highest-ranked
applicable action the decoder constructs is executed; a rollout that proposes
an inapplicable action falls back to the next candidate in the beam, and if
none is applicable the rollout fails. Rollouts also end at the goal, at a
dead end, or at the step bound.

\emph{Plan quality} compares plan lengths on the instances that both the
policy and the reference planner solve. With $|\pi(t)|$ the policy's plan
length on instance $t$ and $|\rho(t)|$ the reference planner's,
\begin{equation*}
P(\pi) \;=\; \frac{\sum_{t \in \mathrm{solved}(\pi) \cap \mathrm{solved}(\rho)} |\rho(t)|}
                 {\sum_{t \in \mathrm{solved}(\pi) \cap \mathrm{solved}(\rho)} |\pi(t)|} .
\end{equation*}
Higher is better, and $P > 1$ means plans shorter than the reference. Because
$P$ conditions on what a system solves, a system with higher coverage is
scored on harder instances, so $P$ should be read within a column rather than
across columns, and coverage is the primary metric.

Coverage is measured with a revisit monitor, which resamples among the
decoder's applicable proposals when a greedy rollout re-enters a state it has
already visited. The monitor belongs to the execution harness and applies
identically to every system, including the untrained control. That is what
makes the control informative: it measures what the harness alone achieves
when the learned component contributes nothing, and it is the number the
policy's coverage should be read against.

\paragraph{In-domain results.} Each in-domain cell of the ladder columns
averages the seven leave-one-out policies whose training set contains the
domain, each evaluated at its lowest-combined-loss checkpoint.
\gadar{} (single-domain) uses the
suite's settings (batch size 64, all training instances of the domain,
500 epochs).

\subsection{Renaming Invariance}
\label{app:invariance}

Invariance to renaming is a property of the construction rather than
something learned, and it is checked by a test that runs on every change to
the representation. Given a domain and a bijective renaming of its
predicates, action schemas, types, and objects, the test builds the graph for
the original and the renamed task and asserts that the node feature matrix,
the edge feature matrix, and the edge lists are identical. It covers both
typing conventions in the benchmark suite and asserts that a domain declaring
PDDL types and an equivalent domain spelling types as unary predicates
produce the same structure after type compilation.

\subsection{Compute and Runtime}
\label{app:compute}

Each training and evaluation run uses a single GPU. Training one model on seven domains takes several hours; a single-domain model takes between half an hour and two hours, so
covering eight domains with dedicated models costs a comparable total and
yields eight separate models rather than one.

\subsection{Code}
\label{app:code}

The implementation is a single Python package built on PyTorch and PyTorch
Geometric, using PDDLGym to parse domains and problems and Fast Downward
\citep{fast_downward} to generate training plans. Each experiment is one YAML
configuration file naming the training domains, the representation, and the
evaluation domains, and a run is a single invocation of the entry point with
one such file, so the configuration files are a complete record of what was
run. The source, those configuration files, and the analysis scripts will be
released publicly under a permissive license on publication, with a pinned
dependency specification.

\end{document}

%% file: math_commands.tex
\usepackage{amsmath,amsfonts,bm}

\def\eqref#1{equation~\ref{#1}}

\def\1{\bm{1}}

\DeclareMathAlphabet{\mathsfit}{\encodingdefault}{\sfdefault}{m}{sl}
\SetMathAlphabet{\mathsfit}{bold}{\encodingdefault}{\sfdefault}{bx}{n}



%% file: iclr2027_conference.bib
@article{generalized_planning_deep_rl,
  title={Generalized planning with deep reinforcement learning},
  author={Rivlin, Or and Hazan, Tamir and Karpas, Erez},
  journal={arXiv preprint arXiv:2005.02305},
  year={2020}
}

@article{llm_baseline,
  title={A systematic evaluation of the planning and scheduling abilities of the reasoning model o1},
  author={Valmeekam, Karthik and Stechly, Kaya and Gundawar, Atharva and Kambhampati, Subbarao},
  journal={Transactions on Machine Learning Research},
  year={2025}
}

@inproceedings{relational_mdp_learning,
  title={Symbolic network: generalized neural policies for relational MDPs},
  author={Garg, Sankalp and Bajpai, Aniket and others},
  booktitle={International Conference on Machine Learning},
  pages={3397--3407},
  year={2020},
  organization={PMLR}
}

@article{policy_beyond_c2,
  title={Learning General Policies for Classical Planning Domains: Getting Beyond C $ \_2$},
  author={St{\aa}hlberg, Simon and Bonet, Blai and Geffner, Hector},
  journal={arXiv preprint arXiv:2403.11734},
  year={2024}
}

@inproceedings{policy_gradients,
  title={Learning general policies with policy gradient methods},
  author={St{\aa}hlberg, Simon and Bonet, Blai and Geffner, Hector},
  booktitle={Proceedings of the International Conference on Principles of Knowledge Representation and Reasoning},
  volume={19},
  pages={647--657},
  year={2023}
}

@article{sr_drl,
  title={Symbolic relational deep reinforcement learning based on graph neural networks and autoregressive policy decomposition},
  author={Janisch, Jarom{\'\i}r and Pevn{\`y}, Tom{\'a}{\v{s}} and Lis{\`y}, Viliam},
  journal={arXiv preprint arXiv:2009.12462},
  year={2020}
}

@inproceedings{rddl_gnn,
  title={Size independent neural transfer for rddl planning},
  author={Garg, Sankalp and Bajpai, Aniket and others},
  booktitle={Proceedings of the International Conference on Automated Planning and Scheduling},
  volume={29},
  pages={631--636},
  year={2019}
}

@inproceedings{expressive_v1,
  title={Learning general optimal policies with graph neural networks: Expressive power, transparency, and limits},
  author={St{\aa}hlberg, Simon and Bonet, Blai and Geffner, Hector},
  booktitle={Proceedings of the International Conference on Automated Planning and Scheduling},
  volume={32},
  pages={629--637},
  year={2022}
}

@inproceedings{goose_v2,
  title={Learning Domain-Independent Heuristics for Grounded and Lifted Planning},
  author={Chen, Dillon Z and Thi{\'e}baux, Sylvie and Trevizan, Felipe},
  booktitle={Proceedings of the AAAI Conference on Artificial Intelligence},
  volume={38},
  pages={20078--20086},
  year={2024}
}

@misc{halo_orchestrator,
      title={Training the Orchestrator: A Supervised Approach to End-to-End PDDL Planning with LLM Agents}, 
      author={Rajesh Mangannavar and Zachary Coalson and Pranay Dugar and Prasad Tadepalli},
      year={2026},
      eprint={2606.21740},
      archivePrefix={arXiv},
      primaryClass={cs.AI},
      url={https://arxiv.org/abs/2606.21740}, 
}

@misc{mueller_qvalues,
      title={Per-Domain Generalizing Policies: On Learning Efficient and Robust Q-Value Functions (Extended Version with Technical Appendix)}, 
      author={Nicola J. Müller and Moritz Oster and Isabel Valera and Jörg Hoffmann and Timo P. Gros},
      year={2026},
      eprint={2603.17544},
      archivePrefix={arXiv},
      primaryClass={cs.AI},
      url={https://arxiv.org/abs/2603.17544}, 
}

@misc{lookahead_policies_2026,
      title={Efficient Lookahead Encoding and Abstracted Width for Learning General Policies in Classical Planning}, 
      author={Michael Aichmüller and Simon Ståhlberg and Martin Funkquist and Hector Geffner},
      year={2026},
      eprint={2605.18674},
      archivePrefix={arXiv},
      primaryClass={cs.AI},
      url={https://arxiv.org/abs/2605.18674}, 
}

@article{mangannavar2026scalable,
  title={Scalable planning through abstraction and learning in fully and partially observable domains},
  author={Mangannavar, Rajesh D},
  year={2026}
}

@misc{hoo_pomdp,
      title={Hierarchical Object-Oriented POMDP Planning for Object Rearrangement}, 
      author={Rajesh Mangannavar and Alan Fern and Prasad Tadepalli},
      year={2025},
      eprint={2412.01348},
      archivePrefix={arXiv},
      primaryClass={cs.LG},
      url={https://arxiv.org/abs/2412.01348}, 
}

@misc{llm_evolved_heuristics_2026,
      title={LLM-Evolved Domain-Independent Heuristics for Symbolic AI Planning}, 
      author={Elliot Gestrin and Jendrik Seipp},
      year={2026},
      eprint={2605.29649},
      archivePrefix={arXiv},
      primaryClass={cs.AI},
      url={https://arxiv.org/abs/2605.29649}, 
}

@misc{learning_to_search_2026,
      title={Learning to Search and Searching to Learn for Generalization in Planning}, 
      author={Michael Aichmüller and Yannik Hesse and Hector Geffner},
      year={2026},
      eprint={2605.25720},
      archivePrefix={arXiv},
      primaryClass={cs.AI},
      url={https://arxiv.org/abs/2605.25720}, 
}

@article{gabar,
  title={Graph Neural Network Based Action Ranking for Planning},
  author={Mangannavar, Rajesh and Lee, Stefan and Fern, Alan and Tadepalli, Prasad},
  journal={Advances in Neural Information Processing Systems},
  volume={38},
  pages={172375--172398},
  year={2025}
}

@inproceedings{wl_goose,
  title={Return to tradition: Learning reliable heuristics with classical machine learning},
  author={Chen, Dillon Z and Trevizan, Felipe and Thi{\'e}baux, Sylvie},
  booktitle={Proceedings of the International Conference on Automated Planning and Scheduling},
  volume={34},
  pages={68--76},
  year={2024}
}

@inproceedings{goose_v1,
  title={GOOSE: Learning domain-independent heuristics},
  author={Chen, Dillon Ze and Thi{\'e}baux, Sylvie and Trevizan, Felipe},
  booktitle={NeurIPS 2023 Workshop on Generalization in Planning},
  year={2023}
}

@inproceedings{gbfs_v2,
  title={Guiding GBFS through learned pairwise rankings},
  author={Hao, Mingyu and Trevizan, Felipe and Thi{\'e}baux, Sylvie and Ferber, Patrick and Hoffmann, J{\"o}rg},
  booktitle={Thirty-Third International Joint Conference on Artificial Intelligence $\{$IJCAI-24$\}$},
  pages={6724--6732},
  year={2024},
  organization={International Joint Conferences on Artificial Intelligence Organization}
}

@article{chrestien_optimize,
  title={Optimize planning heuristics to rank, not to estimate cost-to-goal},
  author={Chrestien, Leah and Edelkamp, Stefan and Komenda, Antonin and Pevny, Tomas},
  journal={Advances in Neural Information Processing Systems},
  volume={36},
  year={2024}
}

@article{ranking_og,
  title={Learning to rank for synthesizing planning heuristics},
  author={Garrett, Caelan Reed and Kaelbling, Leslie Pack and Lozano-P{\'e}rez, Tom{\'a}s},
  journal={arXiv preprint arXiv:1608.01302},
  year={2016}
}

@article{unsupervised_gnn,
  title={Learning generalized policies without supervision using gnns},
  author={St{\aa}hlberg, Simon and Bonet, Blai and Geffner, Hector},
  journal={arXiv preprint arXiv:2205.06002},
  year={2022}
}

@inproceedings{gammazero,
  title={GammaZero: Learning to Guide Belief-Space Search for Long-Horizon POMDPs with Generalizable Graph Representations},
  author={Mangannavar, Rajesh and Tadepalli, Prasad},
  booktitle={Proceedings of the International Conference on Automated Planning and Scheduling},
  volume={36},
  pages={658--666},
  year={2026}
}

@article{complexity-planning,
title = {Complexity, decidability and undecidability results for domain-independent planning},
journal = {Artificial Intelligence},
volume = {76},
number = {1},
pages = {75-88},
year = {1995},
note = {Planning and Scheduling},
issn = {0004-3702},
doi = {https://doi.org/10.1016/0004-3702(94)00080-K},
url = {https://www.sciencedirect.com/science/article/pii/000437029400080K},
author = {Kutluhan Erol and Dana S. Nau and V.S. Subrahmanian},
}

@Misc{pddl-generators,
  author =       "Jendrik Seipp and {\'A}lvaro Torralba and J{\"o}rg Hoffmann",
  title =        "{PDDL} Generators",
  publisher =    "Zenodo",
  year =         "2022",
  howpublished = "\url{https://doi.org/10.5281/zenodo.6382173}"
}

@article{toyer2020asnets,
  title={Asnets: Deep learning for generalised planning},
  author={Toyer, Sam and Thi{\'e}baux, Sylvie and Trevizan, Felipe and Xie, Lexing},
  journal={Journal of Artificial Intelligence Research},
  volume={68},
  pages={1--68},
  year={2020}
}

@inproceedings{strips_hgn,
  title={Learning domain-independent planning heuristics with hypergraph networks},
  author={Shen, William and Trevizan, Felipe and Thi{\'e}baux, Sylvie},
  booktitle={Proceedings of the International Conference on Automated Planning and Scheduling},
  volume={30},
  pages={574--584},
  year={2020}
}

@article{ranking_old_2009,
  title={Learning Linear Ranking Functions for Beam Search with Application to Planning.},
  author={Xu, Yuehua and Fern, Alan and Yoon, Sungwook},
  journal={Journal of Machine Learning Research},
  volume={10},
  number={7},
  year={2009}
}

@article{relational_drl,
  title={Relational deep reinforcement learning},
  author={Zambaldi, Vinicius and Raposo, David and Santoro, Adam and Bapst, Victor and Li, Yujia and Babuschkin, Igor and Tuyls, Karl and Reichert, David and Lillicrap, Timothy and Lockhart, Edward and others},
  journal={arXiv preprint arXiv:1806.01830},
  year={2018}
}

@article{approx_Policy_iter,
  title={Approximate policy iteration with a policy language bias},
  author={Fern, Alan and Yoon, Sungwook and Givan, Robert},
  journal={Advances in neural information processing systems},
  volume={16},
  year={2003}
}

@article{relational_rl,
  title={Relational reinforcement learning},
  author={D{\v{z}}eroski, Sa{\v{s}}o and De Raedt, Luc and Driessens, Kurt},
  journal={Machine learning},
  volume={43},
  pages={7--52},
  year={2001},
  publisher={Springer}
}

@article{fast_downward,
  title={The fast downward planning system},
  author={Helmert, Malte},
  journal={Journal of Artificial Intelligence Research},
  volume={26},
  pages={191--246},
  year={2006}
}

@inproceedings{karia2021learning,
  title={Learning generalized relational heuristic networks for model-agnostic planning},
  author={Karia, Rushang and Srivastava, Siddharth},
  booktitle={Proceedings of the AAAI Conference on Artificial Intelligence},
  volume={35},
  pages={8064--8073},
  year={2021}
}
